\documentclass[a4paper,twoside]{article}

\usepackage{epsfig}
\usepackage{subcaption}
\usepackage{calc}
\usepackage{amssymb}
\usepackage{amstext}
\usepackage{amsmath}
\usepackage{amsthm}
\usepackage{multicol}
\usepackage{pslatex}
\usepackage{apalike}
\usepackage{float}
\usepackage{textcomp}
\usepackage{hyperref}
\usepackage{todonotes}
\usepackage{amsmath}
\usepackage{SCITEPRESS}    
\usepackage{arydshln}

\begin{document}

\title{Multi-Attribute Relation Extraction (MARE) - Simplifying the Application of Relation Extraction}

\author{\authorname{Lars Klöser\sup{1}, Philipp Kohl \sup{1}, Bodo Kraft \sup{1} and Albert Zündorf\sup{2}}
\affiliation{\sup{1}FH Aachen, University of Applied Sciences, Germany}
\affiliation{\sup{2}University of Kassel, Germany}
\email{\{kloeser, p.kohl, kraft\}@fh-aachen.de, zuendorf@uni-kassel.de}
}


\abstract{Natural language understanding's relation extraction makes innovative and encouraging novel business concepts possible and facilitates new digitilized decision-making processes. 
Current approaches allow the extraction of relations with a fixed number of entities as attributes. Extracting relations with an arbitrary amount of attributes requires complex systems and costly relation-trigger annotations to assist these systems.
We introduce \textbf{m}ulti-\textbf{a}ttribute \textbf{r}elation \textbf{e}xtraction (MARE) as an assumption-less problem formulation with two approaches, facilitating an explicit mapping from business use cases to the data annotations. Avoiding elaborated annotation constraints simplifies the application of relation extraction approaches.
The evaluation compares our models to current state-of-the-art event extraction and binary relation extraction methods.
Our approaches show improvement compared to these on the extraction of general multi-attribute relations.}

\onecolumn \maketitle \normalsize \setcounter{footnote}{0} \vfill

\section{\uppercase{Introduction}}
\label{sec:introduction}
Small and medium-sized enterprises (SMEs) increasingly recognize the potential of \textit{natural language understanding} and \textit{relation extraction} to digitalize processes and develop novel software products.
Many product visions include the extraction of variable-sized sets of concept mentions as relations from texts where existing data models define a set of potential attributes per relation.
But most current approaches focus on the extraction of binary relations.  E.g., the number of many thousand biomedical scientific publications per week yielded the successful automation of knowledge discovery \cite{genia-ws-2011-bionlp-shared,genia-kim-etal-2011-overview,genia-kim-etal-2011-overview-genia}. In contrast to many other domains, binarity's structural constraint seems to be reasonable due to cause-effect-relationships. The extraction of more complex semantic relations currently requires the construction of sophisticated systems based on binary classifications. The field of event extraction covers such approaches. Events are multi-attribute relations with so-called trigger annotations. For example, in the following message about a traffic obstruction: \textit{A1 between Köln-Mühlheim and Köln-Dellbrück objects on the road, both directions closed}. According to \cite{linguistic_data_consortium_2005} \textit{closed} triggers the event but provides no event-specific information. Attributes assigned to such triggers build event relations. The central role of trigger annotation results in high-quality requirements and an increased annotation effort. 

This research introduces multi-attribute relation extraction (MARE), a novel problem definition that aims to simplify the application relation extraction approaches in practice. \textit{Multi-attribute relations}: 
\begin{itemize}
\item have a well-known set of potential roles for attributes,
\item make no assumptions on attributes' multiplicity building a relation instance, and
\item do not rely on the trigger concept, indicating one's relation presence.
\end{itemize} 

We introduce a sequence tagging and a span labeling approach to recognize entities and extract multi-attribute relations between them in a joint model. We analyze our approaches' performance on the \textit{SmartData} corpus \cite{schiersch_german_nodate}. This corpus is the only available resource for relation extraction on German texts. This corpus's annotations include named entities and multi-attribute relations between those. We publish all data and source code connected to the research in a GitHub repository\footnote{\url{https://github.com/MSLars/mare}}. Our main contributions can be summarized as follows: 

\begin{itemize} 
\item We formalize multi-attribute relation extraction and introduce two problem-specific approaches.
\item We show that the non-trigger-based approaches have in general a better performance on the multi-attribute relations in the SmartData corpus. 
\item We provide the first reproducible evaluation of a non-binary relation extraction approach on a German corpus. 
\end{itemize} 

\section{\uppercase{Related Work}}
\label{sec:related_work}

Relation extraction investigates the mutual relation between named entities in texts to transfer the unstructured information into predefined schemas. Most benchmark datasets consider only binary relations \cite{mintz_distant_2009,hendrickx_semeval-2010_2010}. 

Traditional \textit{binary relation extraction} approaches use \textit{part of speech tagging}, \textit{dependency parsing}, and further steps to calculate input representations for machine learning models \cite{xu_domain_nodate}. Current state-of-the-art models use transformer networks to calculate highly contextualized representations of binary relation candidates and combine these with specialized decision layers in a combined neural network \cite{li_downstream_2020,eberts_span-based_2019}.

As an extension to binary relation extraction, the field of n-ary relation extraction aims to detect relations with a fixed number of $n$ arguments.  \cite{peng_cross-sentence_2017} extended \textit{recurrent neural networks} to efficiently include syntactic dependency edges to build contextualized relation representations. \cite{lai_bert-gt_2021} presented a transformer-based approach for the same experimental setup. Both focus on 3-ary relations.

Binary and small n-ary relation extraction approaches often enumerate sets of predicted entities as a root for building relation candidates. We extract relations with an arbitrary number of attributes avoiding such an enumeration to prevent a combinatorial explosion.

To extend the fixed-size constraint, the field of \textit{event extraction} defines events as multi-attribute relations with one necessary trigger attribute. The trigger indicates the presence of an event. Other entities can be assigned to single triggers to form event relations \cite{linguistic_data_consortium_2005,aguilar_comparison_2014}. Event extraction approaches rely on this trigger annotations \cite{xiang_survey_2019}. All entities assigned to one trigger form a multi-attribute relation. This reduces the problem to a sequence of binary relation classifications. 

Traditionally, real-world relation extraction systems extract entities and their relationships in a processing pipeline. Such systems suffer from error propagation. The field of \textit{joint relation extraction} investigates models that extract entities and relations in a single model. A common way to build a joint model is to share the embedding layer across multiple downstream tasks. \cite{wadden_entity_2019} introduced a system that shares embeddings to extract named entities, builds binary relation candidates, and classifies the relation between those. \cite{zheng_joint_2017,liu_joint_2019} introduce sequence label schemes to explicitly extract attributes and their relations in a single classification step. Our models similarly extract more complex structures without an enumeration of relation candidates. This avoids a combinatorial explosion for MARE. We apply novel transformer network to receive contextualized text-embeddings \cite{devlin_bert_2019,clark_electra_2020}. 

\cite{schiersch_german_nodate} introduced the SmartData Corpus for relation extraction on German texts. The annotated relations contain a various number of mandatory and optional arguments. Section \ref{sec:data_analysis} analyzes the corpus in great detail. The original paper includes results of the relation extraction system DARE \cite{xu_domain_nodate}. This evaluation considers only mandatory attribute roles for each relation.
We also consider optional attributes and analyze results on a more sophisticated problem setting. \cite{roller_detecting_2018} investigates the extraction of named entities and binary relations from German clinical reports. Their corpus is unpublished.

\section{\uppercase{Data Analysis}}
\label{sec:data_analysis}

We train and evaluate our approaches on the \textit{SmartData}\footnote{\url{https://github.com/DFKI-NLP/smartdata-corpus}} corpus \cite{schiersch_german_nodate}, a German corpus provided by the DFKI\footnote{Deutsches Forschungszentrum für Künstliche Intelligenz (Translation: German Research Center for Artificial Intelligence)}. The corpus contains manually annotated traffic and industry entities and relations in News, RSS feeds, and tweets.

\begin{table}[h]
    \centering
    \caption{The SmartData corpus' train-test-split with the number of relations and the fraction of documents in each subset.}
    \begin{tabular}{c|ccc}
         Data set & Documents & Relations & Ratio \\\hline\rule{0pt}{11pt}
         Training & 1,864 & 1,007 & 0.8\\[3pt]
         Validation & 228 & 129 & 0.1\\[3pt]
         Test & 230 & 128 & 0.1\\\hline\rule{0pt}{11pt}
         Sum & 2,322 & 1,264 & 1\\[1pt]
    \end{tabular}
    
    \label{tab:train-test-split}
\end{table}

The third corpus version contains 19,116 entities and 1,264 relations in 2,322 documents with 141,344 words in total\footnote{Numbers differ from the original paper due to different versions}. \autoref{tab:train-test-split} shows the train-test-split provided by SmartData.

The \textit{inter-annotator agreement} is on a moderate level \cite{viera2005understanding} with Cohen's kappa coefficient for entities of 0.58 and 0.51 for relations.
DFKI describes their preprocessing steps in \cite{schiersch_german_nodate} and GitHub.
\begin{figure}[h]
    \centering
    \includegraphics[scale=0.55]{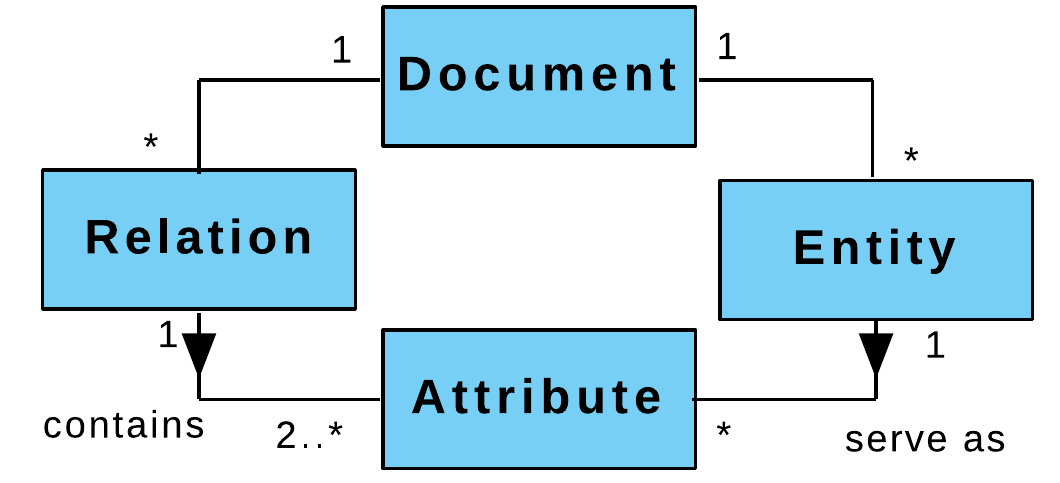}
    \caption{Structure of instances in the SmartData corpus. Documents contain relations and entities. A relation has at least two mandatory attributes. Each attribute has an entity mention. Entities may function as attributes in zero or multiple relations. E.g., the entity \textit{Location} may serve as an attribute in \textit{Accident} and \textit{Obstruction} at the same time.}
    \label{fig:data-meta-model}
\end{figure}

\autoref{fig:data-meta-model} illustrates the data meta-model. Note that a relation can have a variable number of attributes and is not limited to a fixed number. For each attribute's role a fixed set of entity types may fit: E.g., entity types such as \textit{Location-Street}, \textit{Location-City} or \textit{Location-Route} can serve interchangeably as an attribute with role \textit{Location}.

In the following, we explain the key characteristics of the SmartData corpus.

\begin{figure*}[h]
    \centering
    \includegraphics[scale=0.54]{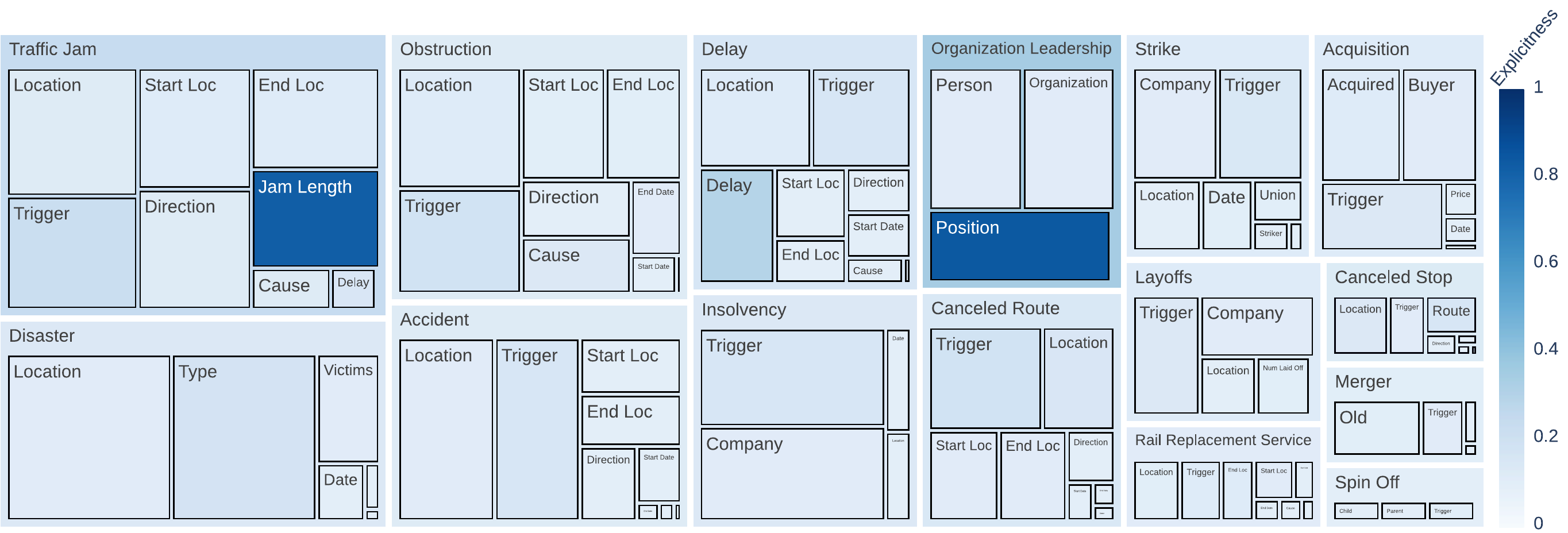}
    \caption{Distribution of relations and attributes. The rectangles' sizes are proportional to the relation or attribute frequency. The explicitness is the quotient of the attributes' frequency and the total number of entities with an entity type suitable for the specific attribute role. The metric indicates how reliably an entity type indicates a relation attribute.}
    \label{fig:rel_arg_distribution}
\end{figure*}

\textbf{Relations.}
The corpus provides 15 relation types with two mandatory and arbitrary optional attributes. \autoref{fig:rel_arg_distribution} illustrates the relation's and the attribute's distribution.

\begin{figure}[h]
    \centering
    \includegraphics[scale=0.48]{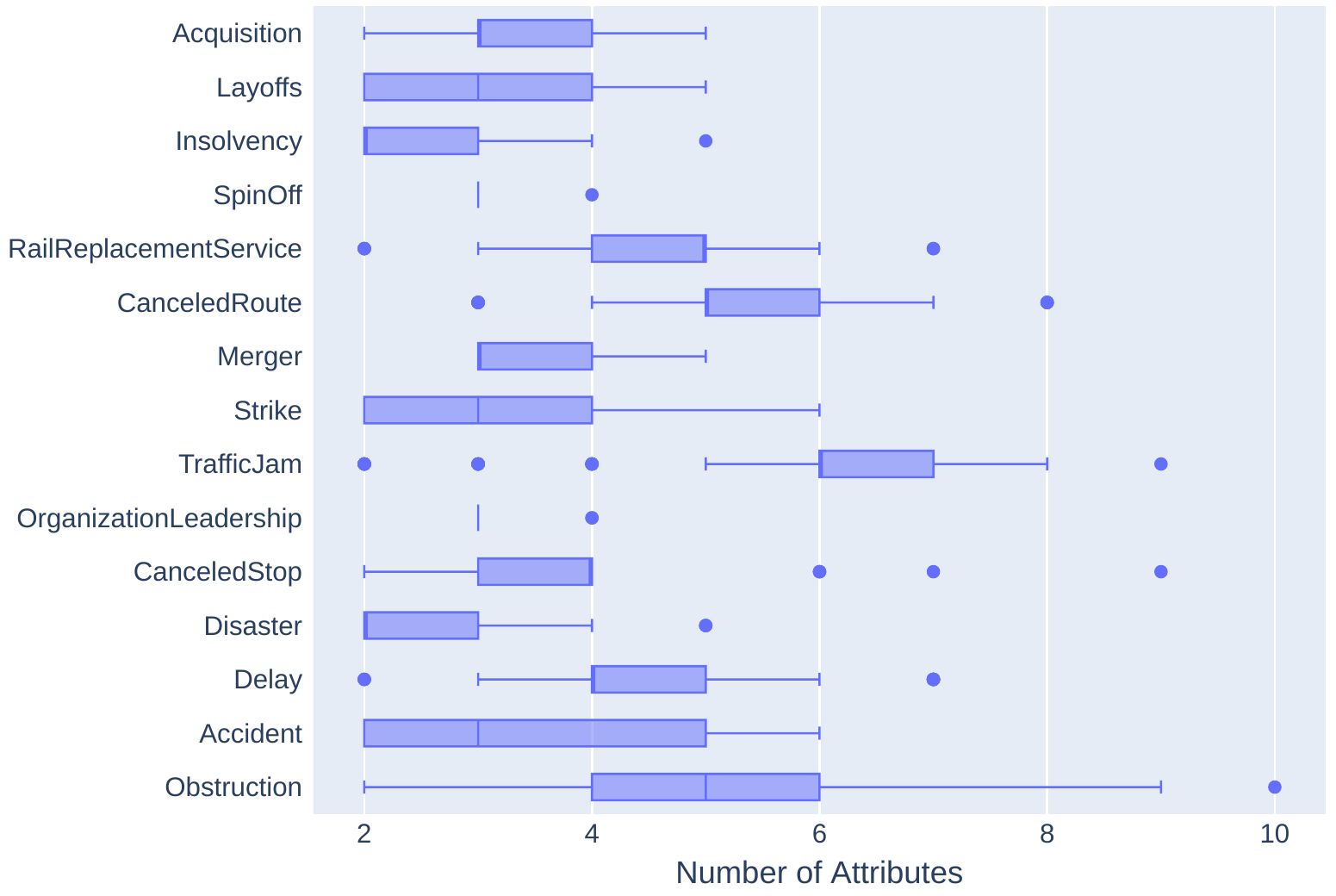}
    \caption{Boxplot illustrating the distribution of the number of attributes per relation. E.g., the number of \textit{Obstruction's} attributes ranges from two to ten while other relations like \textit{Insolvency} do not show the same variance. Dots indicate outliers.}
    \label{fig:arg_num_distribution}
\end{figure}

\textbf{Entities.} SmartData provides 16 fine-grained entity types. For a full list, see \cite{schiersch_german_nodate}. We introduce \textit{explicitness} as metric to demonstrate that only a few entity types are a strong indicator for a relation (cf. \textit{Jam Length} or \textit{Position} in \autoref{fig:rel_arg_distribution}). Therefrom, MARE models have to learn a combined view of entity compounds.

\textbf{Variable Number of Relation Attributes.} 
Each relation contains at least one of each mandatory attributes. They may or may not contain further optional attributes. The example for \textit{Arguable differences} in \autoref{fig:error-types} shows an RSS feed with an \textit{Obstruction} relation. Only \textit{trigger} and \textit{location} attributes are mandatory. \textit{StartLoc} and \textit{EndLoc} are optional attributes.

\textbf{Unbalanced.} Unbalanced datasets raise the challenge to learn the essential structure for underrepresented data points and the other richer data points \cite{unbalanced_dataset}. The dataset is unbalanced in terms of relations as of attributes as well (cf. \autoref{fig:rel_arg_distribution}): \textit{Traffic Jam} occurs approximately 10 times more often than \textit{Spin Off}. While Spin Off's attributes frequencies are quite equal, Traffic Jam's attributes show a difference between the attribute's distribution, which corresponds to mandatory and optional attributes.

\textbf{Improper Triggers.} Other event extraction corpora's triggers are strictly defined to one single mandatory token or span due to its essential role as relation indicator \cite{linguistic_data_consortium_2005,aguilar_comparison_2014}. SmartData does not follow these constraints: the triggers are optional and not bound to consecutive tokens or any specific lemma or part of speech. Thus, this corpus impedes the application of current event extraction approaches due to their assumption of one existing trigger token/span.

\textbf{Relations Share Entities.} Multiple relations can occur in one document. The corresponding relation's entities dot not have to be disjointed: e.g., \textit{Traffic Jam} and \textit{Obstruction} likely appear together and sharing \textit{location} attributes. 

\textbf{Different Register of Language.} SmartData uses different data sources leading to different distributions, and patterns models have to learn. While news articles are continuous and grammatically valuable text, Twitter and RSS feeds are often sentence fragments.

The SmartData corpus provides relations with a variable amount of attributes and without a regular trigger definition, making the corpus fit into the MARE definition. Modifications are necessary to apply current relation or event extraction approaches.

\section{\uppercase{MARE}}
\label{sec:experiments}

This section formally introduces the concept of multi-attribute relation extraction and introduces two MARE approaches. We describe our evaluation methodology, which includes the adaptation of an event and binary relation extraction approach. We compare both against the MARE approaches.

\subsection{Definition}
\label{sec:problem_definition}

For a given text $t = (t_1, \dots, t_n)$ with $n$ tokens,
$$
S = \{(t_i, \dots, t_j) \;| \mbox{ for all } i,j \in \{1, \dots, n\}, i \leq j\}
$$
denotes the set of all text-spans. Let $L$ be a set of relation labels and $A_l$ be a set of attribute roles for each relation label $l \in L$. The task is to predict a relation set $R$ for a given text $t$. Each relation instance $r \in R$ 
$$
r = (l, \{\alpha_i \;| \mbox{ for all } i \in \{1 ,\dots, m \}\})
$$
consists of a relation label $l$ and a variable number of $0 < m \le |S|$ attributes
$$\alpha_i = (s, a) \in S \times A_l \mbox{ for all } i \in \{1, \dots, m\}.$$

Each span $s \in S$ can contribute to at most one attribute in each relation $r \in R$. But a span can contribute to attributes in multiple relations. We explicitly allow relations with one attribute. We denote text-spans $s_{ij}$ with $i,j$ as start and end indices. Further, $$A = \bigcup_{l\in L} A_l$$ is the set of all attribute roles.

The formal definition makes no difference between mandatory and optional attributes as in \autoref{sec:data_analysis}. We still use this distinction for the model evaluation since a higher frequency of an attribute role implies a better extraction performance.

\subsection{Approaches}
\label{sec:approaches}

All approaches except the baselines use transformer networks as contextualized embedders. Such networks compute contextualized representations with a combination of multiple \textit{self-attention} and \textit{feed-forward-layers}. They are trained in an unsupervised fashion \cite{devlin_bert_2019}. We apply a german version of ELECTRA\footnote{\url{https://huggingface.co/german-nlp-group/electra-base-german-uncased}} \cite{german-nlp-group2020}. Its pretraining tasks focus on the models' ability to describe the semantic structure of texts \cite{clark_electra_2020}. All approaches use the version of the Adam optimization algorithm with weight decay introduced in \cite{loshchilov2019decoupled}.

In the following, we use the definitions introduced in \autoref{sec:problem_definition}.

\subsubsection{Sequence Tagging}
\label{sec:sequence_tagging}
The unknown number of attributes in MARE requires models that do not need to enumerate all relation candidates.
\cite{zheng_joint_2017} introduced a tagging scheme to formulate binary relation extraction as a sequence tagging problem.
$$
T = \{b, i\} \times L \times A \cup \{o\}
$$
describes our tag-set. Tags that start with $b$ and $i$ mark tokens as the beginning or inner part of an entity. For the resulting entity spans, the label $l \in L$ determines the relation, and $a \in A_l$ determines the attribute role for a relation determined by $l$. $o$ marks tokens that do not belong to an attribute.

From each tagged token sequence, we extract a set of incoherent relation attributes. These attributes are summarized to relation instances by their relation label.

The embedded and contextualized input sequence is the input for a feed-forward layer, which maps them to label probabilities. A \textit{conditional random field} determines the loss and the most probable label sequence. \cite{huang_bidirectional_2015} describes the details of conditional random fields for sequence tagging.

Our sequence tagging model avoids the enumeration of all potential relation candidates. However, this also leads to the two following limitations:

\begin{enumerate}
    \item \textbf{Shared attributes across relations}. Multiple relations may have attributes with shared text spans. Our tagging scheme can only assign each span to at most one relation.
    \item \textbf{Multiple relations with the same label}. A sample could have multiple relations with the same relation label. For example, two accident descriptions in one sample. A grouping based on the label leads to a single relation instead of multiple instances.
\end{enumerate}

We introduce a layer of business logic to deal with such situations. In the case of shared attribute spans across various relations, we check whether the current relations have any missing mandatory attributes.  If so, we search for attribute types indicating shared arguments. If such an attribute is within a \textit{maximum relation width}\footnote{The maximum relation width is a hyperparameter and determined by a hyperparameter search. The GitHub repository contains the search configuration and final values.}, we use it to complete the relation.  

In the following, we assume relation attributes to be sorted by their span indices. To handle multiple relations with the same label in one sample, we split a grouped relation $\alpha_1, ..., \alpha_n$ at an index $i < n$ if the subsets $\alpha_1, \dots, \alpha_i$ and $\alpha_{i+1}, \dots ,\alpha_n$ contain all mandatory attributes and the distance between $\alpha_i$ and $\alpha_{i+1}$ exceeds the maximum relation width.

\subsubsection{Span Labeling}
\label{sec:span_labeling}

Our second approach is motivated by \cite{liu_joint_2019}. They applied a sequence labeling approach instead of sequence tagging. Labeling allows the assignment of multiple attribute labels to each text-span. 
We modify this approach and predict a relation-attribute label for each possible text-span in a given sample. As our sequence tagging approach, this approach does not need to enumerate all relation candidates and resolves the limitation of shared attributes across relations.

\begin{figure}[h]
    \centering
    \includegraphics[scale=0.28]{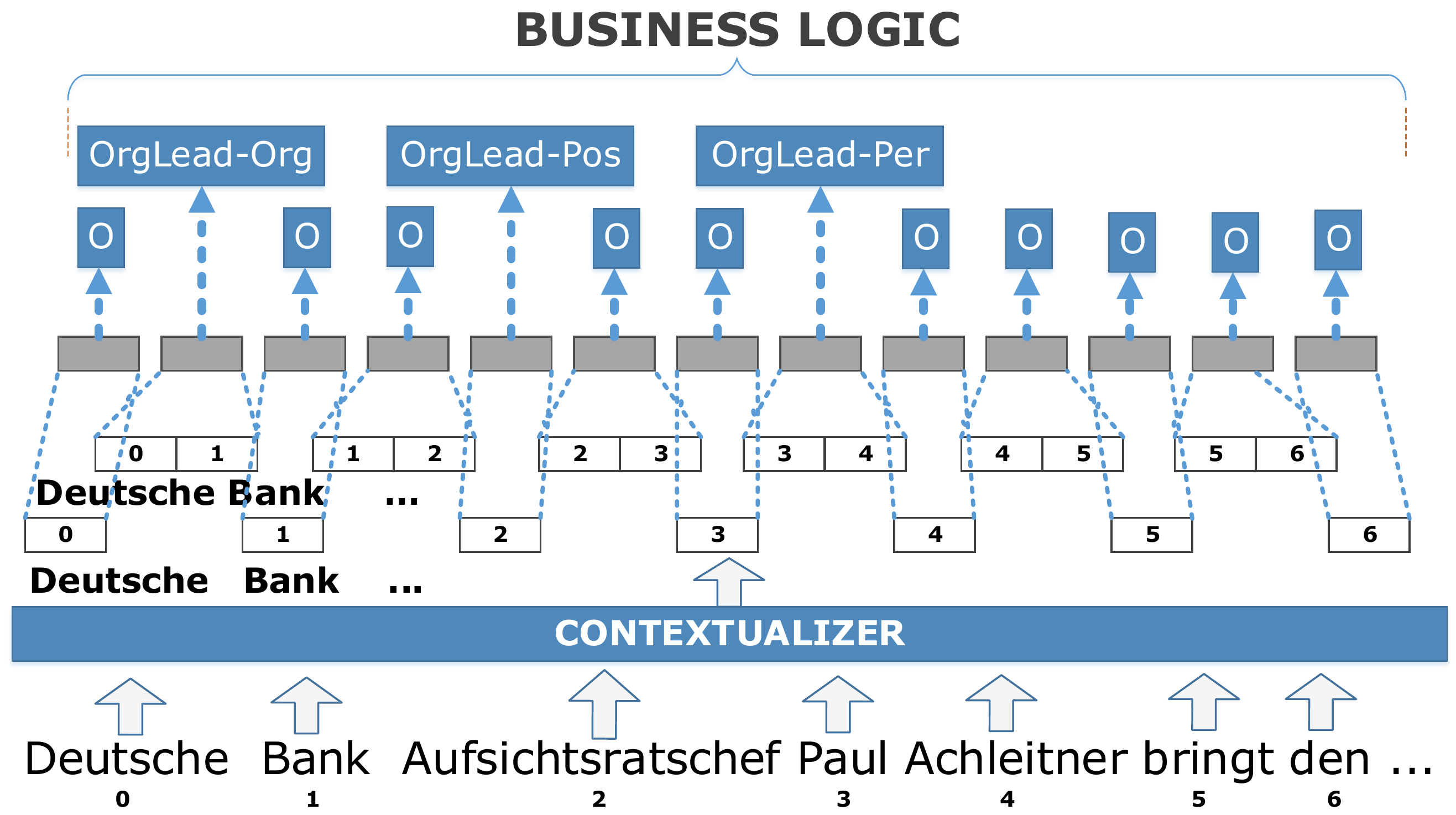}
    \caption{Illustration of the span labeling approach. The input sequence is embedded and contextualized. Each text-span within a \textit{maximum span width} (2 in this example) is transformed to a fixed-length representation labeled with a relation-label and argument-role combination. Finally, the business logic groups attribute to relations.}
    \label{fig:span_spproach}
\end{figure}

Let $T =  L \times A$ be a set of labels indicating the relation label and attribute role for a given text-span. The model predicts a probability $P(t | s)$ for each label $t \in T$ and each span $s \in S$. A \textit{maximum span width} hyperparameter defines the maximum number of tokens per span in $S$. We apply a \textit{binary cross-entropy loss} function, which allows the assignment of multiple labels per span.

\autoref{fig:span_spproach} illustrates our model architecture. 
The span representations are computed with a self-attention-based module from AllenNLP\footnote{\url{http://docs.allennlp.org/main/api/modules/span_extractors/self_attentive_span_extractor/}}. For a given text with length $n$ we compute contextualized embeddings $(c_1, \dots, c_n)$ of dimension $d$. 
For each span $s_{ij}$, we have $j-i+1$ embeddings $(c_i, \dots, c_j)$. To get a fixed-length span representation of dimension $d$, we calculate a linear combination of these embeddings. A parameter matrix $M \in \mathbb{R}^{d \times 1}$ calculates global attention scores
$
a_i = c_i \cdot M
$
for all  $i \in \{1,\dots,n\}$. These are used to calculate weigths $w_i, \dots, w_j$ for a span $s_{ij}$, with
$$
w_k = \frac{e^{a_k}}{\sum_{i \leq l \leq j} e^{a_l}} \mbox{ for all } k \in \{i, \dots, j\}.
$$
The softmax function ensures that the weights for each span sum-up to 1.
The final span representations
are a linear combination of these weights and the embeddings.

A \textit{feed forward layer} in combination with the element-wise \textit{sigmoid} function computes the label probabilities for each span. Similar to the previous approach, this leads to a set of grouped relation instances. We apply the same business logic as in \autoref{sec:sequence_tagging} since the limitation of \textit{multiple relations with the same label} remains.

\subsubsection{Event Extraction}

To apply event extraction approaches, we need to specify an event trigger for each multi-attribute relation.
As \autoref{sec:data_analysis} shows, some instances in the SmartData corpus have no such annotations. If a relation definition has no mandatory trigger attribute, we defined one mandatory attribute type for each relation as the trigger. In the case of multiple and non-conjunct trigger spans, we select the first span as the trigger. We did not apply more complex logic since the set of relations with multiple triggers (78 of 1264) is relatively small. The first error situation in \autoref{sec:sequence_tagging} is unsolved if relations share triggers. Other attributes can be shared across relations.

We apply Dygie$++$\footnote{\url{https://github.com/dwadden/dygiepp}} as event extraction approach. As \cite{wadden_entity_2019} describes, Dygie$++$ uses contextualized span representations, similar to \autoref{sec:span_labeling}. Trigger detection and attribute disambiguation use this shared span representations. 

\subsubsection{Binary Relation Extraction}

Many binary relation extraction approaches classify all possible pairs of entities as relation candidates, as SpERT  \cite{eberts_span-based_2019}. 
In combination with multi-class labeling, this solves both error situations in  \autoref{sec:sequence_tagging}.

We apply SpERT to extract binary relations from 1,717 of 1,864 samples in the train split that contain relations with exactly two mandatory attributes. The next section introduces various evaluation strategies. We introduce a \textit{binary relation extraction} strategy to compare the performance of SpERT against all other approaches on the subset of valid binary relations.

\subsection{Experimental Setup}
\label{sec:experimental_setup}

We used AllenNLP \cite{Gardner2017AllenNLP} and Pytorch to implement the sequence tagging and span labeling approach. Our GitHub repository contains modified versions of Dygie$++$ and SpERT. These modifications were necessary to integrate both approaches into our experimental infrastructure.

Our GitHub repository contains a summary of all hyperparameters and their values in the final models. All hyperparameters were determined with Optuna\footnote{\url{https://optuna.org/}}. We applied 50 optimization trials for each model. We fixed the learning rate for the transformer network's embedding layer to $5 \cdot 10^{-5}$ and $10^{-3}$ for all other network components. We selected a batch size of 6 for all MARE approaches, for SpERT and Dygie$++$ a batch size of 1.

We use various evaluation strategies to analyze the predictions. The strategies aim to reflect the challenges of MARE on different levels of complexity.

\begin{itemize}
    \item \textbf{Attribute Recognition (AR)} The evaluation is made attribute wise. An attribute is considered correct if its boundaries, relation label, and attribute role are predicted correctly while not considering the grouping to a relation. 
    \item \textbf{Classification (Cl)} A prediction is correct if the predicted label matches a gold relation label.
    \item \textbf{Mandatory Relation Extraction (MRE)} A prediction is correct if all mandatory attributes and the relation label match against the gold annotation. Thus, the grouping of the mandatory attributes to one relation is essential.
    \item \textbf{Complete Relation Extraction (CRE)} Measures the model's capability of extracting the relation with all attributes as a whole. Thus, a prediction is considered correct if the model extracts all attributes and groups them correctly into a relation with the right relation label.
    \item \textbf{Binary Relation Extraction (BRE)} This is the MRE strategy on the subset of samples that contains only relations with exactly two mandatory arguments. This strategy allows a comparison between SpERT and all other approaches.
\end{itemize}

We include the baseline (DARE) from \cite{schiersch_german_nodate}, which focuses on the mandatory arguments and uses gold entity annotations. Our own baseline is a modification of the sequence tagging approach. We replace the pretrained transformer network with a combination of GloVe\footnote{\url{https://deepset.ai/german-word-embeddings}} word vectors \cite{pennington-etal-2014-glove} and character level CNN as embedding layer. A Bi-GRU layer contextualizes the inputs. 

Our computational setup contains two nodes with Intel Xeon Platinum 8168 CPUs, Nvidia Quadro P5000 GPUs with 16 GB RAM, and Ubuntu 18.04 OS. A hyperparameter search took approximately 24 hours.

\section{\uppercase{Results}}
\label{sec:results}

The metrics in \autoref{tab:final_result_values} measure different capabilities necessary to extract all attributes, their roles, and the relation label in combination. In general, as the requirements of the metrics increase, the metrics' values decrease.

The span labeling approach increases the event extraction AR results by $0.06$ and the Cl results by $0.04$ in the F1 scores. We observe that both MARE approaches perform better than Dygie$++$ on the complete dataset. The similar MRE score and the increased AR and CRE scores indicate that MARE models extract optional and potentially less-frequent arguments more reliably. A reduction to the subset of documents with exactly two mandatory arguments leads to higher general scores but a much higher increase for Dygie$++$ than for the MARE models. Both observations indicate that model architectures with fewer structural assumptions are better suitable for the corpus' unique characteristics as described in \autoref{sec:data_analysis}.

\begin{table}[h]
    \centering
    \caption{Model evaluation on the test set based on different strategies, see \autoref{sec:experimental_setup}. Precision, Recall and F1 score serve as comparison metrics.}
    \label{tab:final_result_values}
    \begin{tabular}{ll|cccc:c}
    Model       &       & AR            & Cl            & MRE           & CRE           & BRE           \\\hline \hline
    MARE        & F1    & .66           & .76           & .45           & \textbf{.30}  & .47           \\
    Seq. Tag.   & P     & .66           & .73           & .43           & .28           & .44           \\
                & R     & \textbf{.66}  & .80           & \textbf{.48}  & \textbf{.31}  & .49           \\
    \hline\hline
    MARE        & F1    & \textbf{.70}  & \textbf{.80}  & \textbf{.47}  & .29           & .49           \\
    Span Lab.   & P     & \textbf{.75}  & \textbf{.80}  & .47           & \textbf{.29}  & .49           \\
                & R     & .65           & \textbf{.80}  & .47           & .29           & .49           \\
    \hline\hline
                & F1    & .64           & .76           & .46           & .25           & \textbf{.53}  \\
    Dygie$++$   & P     & .63           & .77           & .47           & .26           & .55  \\
                & R     & .65           & .74           & .45           & .25           & \textbf{.52}           \\
    \hline\hline
                & F1    & -           & -           & -           & -           & .51           \\
    SpERT       & P     & -           & -           & -           & -           & \textbf{.57}           \\
                & R     & -           & -           & -           & -           & .45           \\
    \hline\hline
    MARE        & F1    & .60           & .68           & .39           & .26           & .41           \\
    Baseline    & P     & .66           & .68           & .40           & .26           & .40           \\
                & R     & .55           & .67           & .39           & .26           & .41           \\
    \hline\hline
                & F1    & -             & -             & .28           & -             & -             \\
    DARE        & P     & -             & -             & \textbf{.53}  & -             & -             \\
                & R     & -             & -             & .19           & -             & -             \\
    \end{tabular}
\end{table}

\begin{figure*}
    \centering
    \includegraphics[width=\linewidth]{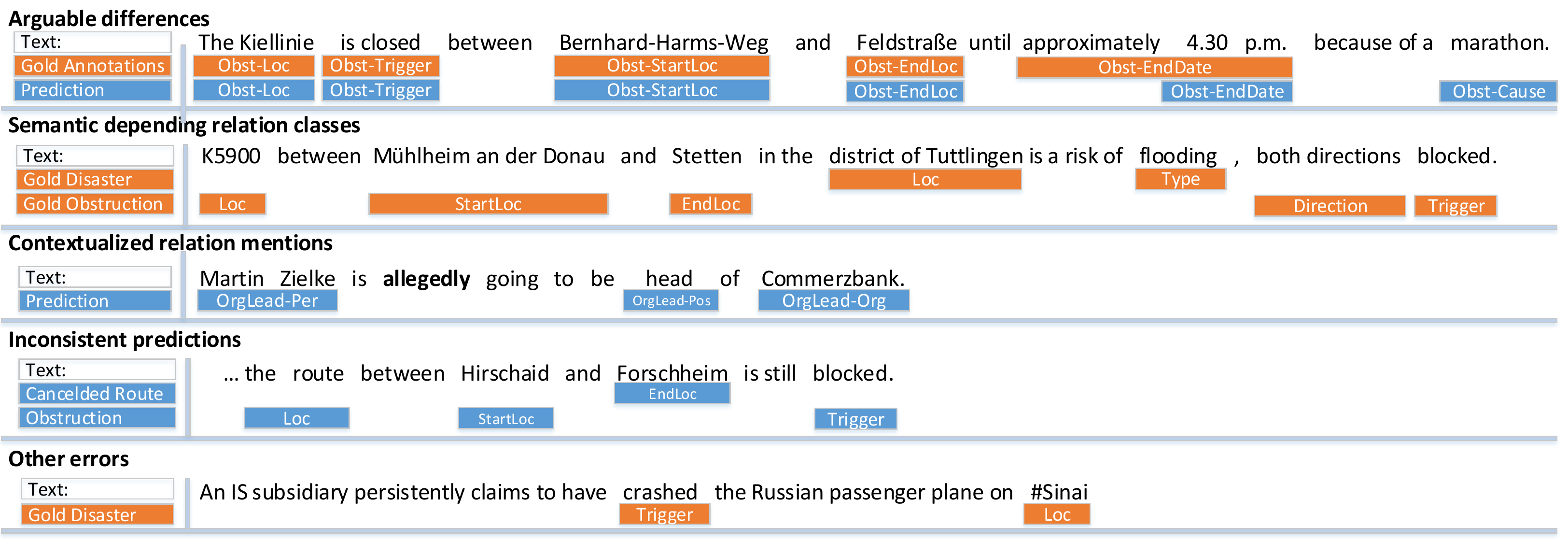}
    \caption{Examples for error classes. Colored boxes indicate the relations and their attributes. The attributes role is textually annotated. Predictions made by the span labeling approach.}
    \label{fig:error-types}
\end{figure*}

The difference between the MARE baseline and both MARE approaches shows the positive effect of pretrained transformer networks. Despite the general weaker performance of DARE, which uses an automatically selected rule-set, the original benchmark has the highest MRE precision score. This indicates a high certainty in extracted relations and a high number of false negatives because of the low recall score.

The evaluation of SpERT shows a clear improvement compared to our MARE baseline. SpERT performs also better than our MARE models on BRE.

\begin{table}[h]
    \centering
    \caption{Comparison of approaches trained with and without trigger annotations. We exclude trigger entities from the score computations.}
    
    \begin{tabular}{l|cc}
         Model                     & AR F1          & MRE F1        \\
         \hline\hline
         Seq. Tag. with Trigger    & .64            & .49               \\
         Seq. Tag. without Trigger & .64            & .51               \\
         \hline\hline
         Span Lab. with Trigger    & .67            & .52               \\
         Span Lab. without Trigger & .66            & .54               \\
         \hline\hline
         Dygie$++$. with Trigger    & .62            & .53               \\
    \end{tabular}
    
    \label{tab:no-trigger}
\end{table}

\autoref{tab:no-trigger} shows how trigger annotations effect MARE models' performance. Evaluations on the reduced set of non-trigger attributes of MARE models trained both with or without trigger annotations do not show a significant difference in AR and MRE scores. Our models' performance without trigger annotations is comparable to state-of-the-art event extraction on multi-attribute relations from the SmartData corpus. The MARE models AR score is better than Dygie$++$'s. That proves MARE's ability to extract optional and less-frequent attributes. 

Compared to \autoref{tab:final_result_values} the AR scores decrease, indicating that the models extract trigger attributes reliably. Without trigger attributes many single-attribute relations remain. This simplifies the MRE task and causes the increased MRE scores.

Since relation extraction is a task on a high semantic level and SmartData's gold annotations contain a certain degree of inconsistency, we provide a manual error analysis to understand our models' prediction characteristics better.

\subsection{Error Inspection}

We conducted a manual comparison of the differences between the gold annotations and the models' predictions. The following error-equivalence-classes emerge from our manual inspection. All examples mentioned in the following enumeration refer to \autoref{fig:error-types}.

\begin{enumerate}
    \item \textbf{Arguable differences} The models' predictions are often reasonable even if the gold data contain divergent annotations. The example shows an \textit{Obstruction} relation, in which the \textit{marathon} represents the \textit{Obstruction-Cause}.
    The gold annotation does not reflect this circumstance. 
    Arguable differences indicate that our models learned certain semantic concepts. Some generalizations in the predictions lead to false positives lowering the evaluation metrics.
    \item \textbf{Semantic depending relation classes} Some relation classes, such as \textit{Accident} and \textit{Obstruction} have a strong semantic relationship. Therefore, instances of these relations are often nested and share entity spans as attributes. The annotations of these shared attributes are often flawed. The example shows an \textit{Obstruction} caused by a \textit{Disaster}. The gold annotation contains two separate relations and does not express this dependency. The trigger of the \textit{Disaster} could also be interpreted as the cause of the \textit{Obstruction}. This distinction is challenging for the models.
    \item \textbf{Contextualized relation mentions} Many supposed relation instances appear in a presuming context. Words like \textit{allegedly} indicate assumptions rather than facts. The example shows a presumption about an \textit{Organization Leadership}. In many of these cases, the models predicted relation instances.
    \item \textbf{Inconsistent predictions} The example shows an \textit{Obstruction} relation, where the model predicted all roles correctly. The relation label for the \textit{End Location} belongs to a similar semantic relation. If the missing attributes are not mandatory, such situations cannot be resolved by the business logic.
    \item \textbf{Other errors} Many relations are not recognized by the models. Often such errors occur in sentences with less grammatical structure and sentences that contain many special characters like '@', '\#' or typical trigger phrases that do not belong to any relation. The example shows a \textit{Disaster} that no model predicted. 
\end{enumerate}

The results indicate that current event or binary relation extraction approaches outperform MARE models on the task of binary relation extraction. However, as we weaken the structural requirements, MARE models become superior. The introduced MARE approaches allow the extraction of complex multi-attribute relations from plain text without an enumeration of all relation candidates. The limitations of our approaches, see \autoref{sec:approaches}, had no severe impact concerning the smart data corpus.

\section{\uppercase{Conclusion}}
\label{sec:conclusion}

We introduced multi-attribute relation extraction and differentiated this definition from current terminology as n-ary relation extraction and event extraction. Our problem definition leads to simplified approaches for extracting relations with an arbitrary amount of attributes by avoiding the usage of candidate enumeration and the trigger concept.

MARE models are superior if the relations do not fit into binary or event schema. They avoid structural constraints and perform better than current state-of-the-art relation and event extraction approaches on the SmartData corpus. 

We plan to involve the manual analysis results in the construction of improved MARE approaches in the future. Mostly we want to address the limitations the MARE approaches have and the incorporation of the relation-specific context.

\bibliographystyle{apalike}
{\small
\bibliography{main}}

\end{document}